\documentclass[conference, 10 pt, conference]{ieeeconf}  % Comment this line out if you need a4paper

\usepackage{amsmath,amsfonts}
\usepackage{algorithm}
\usepackage{array}
\usepackage[caption=false,font=normalsize,labelfont=sf,textfont=sf]{subfig}
\usepackage{amsmath,amssymb}
\usepackage{textcomp}
\usepackage{stfloats}
\usepackage{url}
\usepackage{verbatim}
\usepackage{graphicx}
\usepackage{cite}
\usepackage{bm}
\usepackage{empheq}
\usepackage{mathrsfs}
\usepackage{subcaption}
\usepackage{multirow}
\usepackage{booktabs}
\usepackage{subfig}
\usepackage{algpseudocode}
\usepackage{amsmath}
\usepackage{xcolor}
\usepackage[hidelinks]{hyperref}

\IEEEoverridecommandlockouts                              % This command is only needed if 
\title{\LARGE \bf
Easy-IIL: Reducing Human Operational Burden in Interactive Imitation Learning via an Assistant Expert
}

\author{Chengjie Zhang$^{1,2}$, Chao Tang$^{3}$, Wenlong Dong$^{1,2}$, Dehao Huang$^{1,2}$, Aoxiang Gu$^{1,2}$, \\ and Hong Zhang$^{1,2}$ \emph{Life Fellow, IEEE}% <-this % stops a space
\thanks{$^{1}$Shenzhen Key Laboratory of Robotics and Computer Vision, Southern University of Science and Technology, Shenzhen, China.}%
\thanks{$^{2}$Department of Electronic and Electrical Engineering, Southern University of Science and Technology, Shenzhen, China.}%、
\thanks{$^{3}$Department of Robotics, Perception, and Learning, KTH Royal Institute of Technology, Stockholm, Sweden.}%
\thanks{This work was supported in part by the Shenzhen Key Laboratory of Robotics and Computer Vision (ZDSYS20220330160557001).}
}

\begin{document}

\maketitle
\thispagestyle{empty}
\pagestyle{empty}

%%%%%%%%%%%%%%%%%%%%%%%%%%%%%%%%%%%%%%%%%%%%%%%%%%%%%%%%%%%%%%%%%%%%%%%%%%%%%%%%
\begin{abstract}
Interactive Imitation Learning (IIL) typically relies on extensive human involvement for both offline demonstration and online interaction. Prior work primarily focuses on reducing human effort in passive monitoring rather than active operation. Notably, structured model-based imitation approaches achieve comparable performance with significantly fewer demonstrations than end-to-end imitation learning policies in the low-data regime. However, these methods are typically surpassed by end-to-end policies as the data increases. Leveraging this insight, we propose Easy-IIL, a framework that utilizes off-the-shelf model-based imitation methods as an assistant expert to replace active human operation for the majority of data collection. The human expert only provides a single demonstration to initialize the assistant expert and intervenes in critical states where the task is approaching failure. Furthermore, Easy-IIL can maintain IIL performance by preserving both offline and online data quality. Extensive simulation and real-world experiments demonstrate that Easy-IIL significantly reduces human operational burden while maintaining performance comparable to mainstream IIL baselines. User studies further confirm that Easy-IIL reduces subjective workload on the human expert.
\end{abstract}

%%%%%%%%%%%%%%%%%%%%%%%%%%%%%%%%%%%%%%%%%%%%%%%%%%%%%%%%%%%%%%%%%%%%%%%%%%%%%%%%
\section{INTRODUCTION}\label{sec:intro}
\begin{figure}[t]
\centerline{\includegraphics[width = 0.95\linewidth]{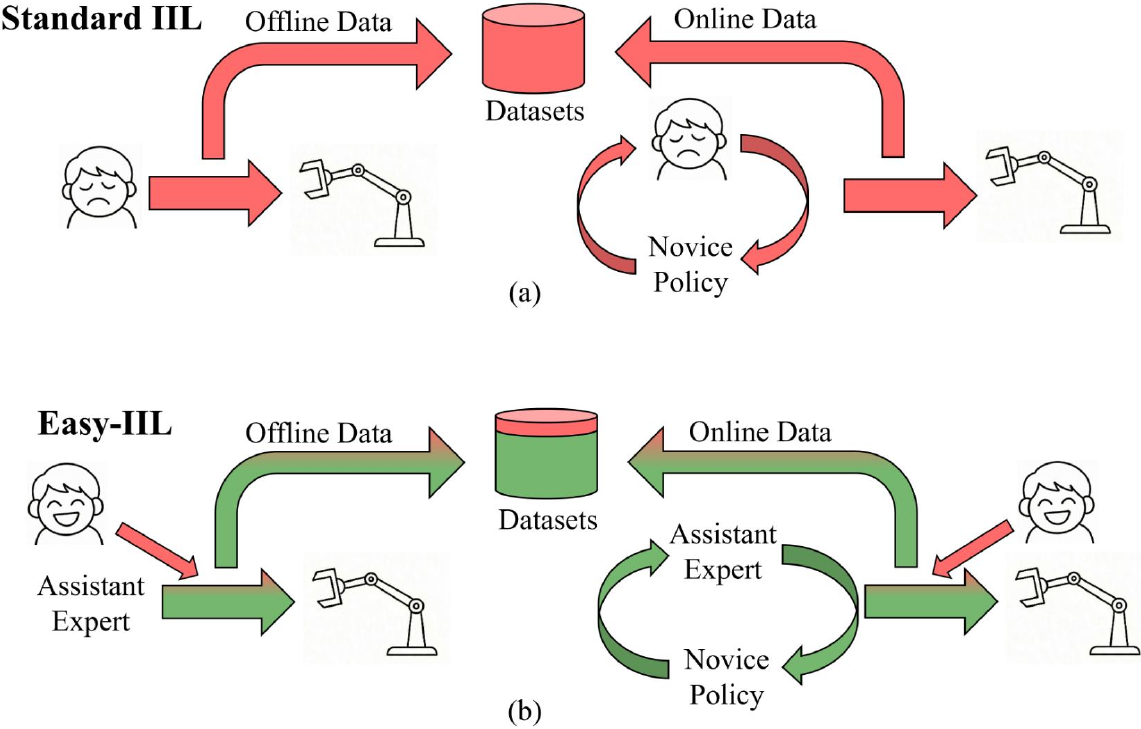}}
\caption{(a) illustrates the standard IIL interaction mechanism, where a human expert is involved in both offline and online stages. (b) depicts Easy-IIL, which delegates most offline and online data collection and interactions to an assistant expert, substantially reducing active human operational effort.}

\label{fig:intro_iil}
\end{figure}

Imitation learning (IL) constitutes a paradigm for acquiring skills through the observation of expert demonstrations~\cite{il_manip}. Building on this, Interactive Imitation Learning (IIL)~\cite{iil_survey} further enhances sample efficiency by introducing online interaction, which allows an expert to correct the compounding errors during policy rollouts. This has been validated across various domains, such as virtual games \cite{dagger}, autonomous driving \cite{hg_dagger}, and robotic manipulation \cite{sirius}. Standard IIL, as shown in Fig.~\ref{fig:intro_iil}(a), trains a novice from an expert-collected (a human in our case) offline dataset and then iterates online rollout, expert intervention at near-failure states, data aggregation, and retraining~\cite{iil_survey}. However, such a framework places a heavy burden on the human expert, limiting scalability as dataset size and task complexity grow and potentially inducing fatigue that degrades the quality of collected data.

% The standard IIL framework \cite{iil_survey} begins with an expert collecting an offline dataset to train a novice policy. This is followed by an iterative procedure consisting of online deployment, expert correction of compounding errors, data aggregation, and retraining on the expanded dataset. Expert correction refers to timely intervention triggered by the detection of near-failure states (including colliding with the environment or leaving the target object in an unrecoverable state). In the context of our study, a human serves as the expert. Fig. \ref{fig:intro_iil}(a) shows the data collection procedure of standard IIL.

To address the above challenge, prior work reduces the burden of \textit{passive} human monitoring by developing real-time anomaly detection to inform when human intervention is needed~\cite{hg_dagger, model_based}. In this work, we push further toward reducing the \textit{active} operational burden, the effort required to teleoperate the robot during both offline and online stages, while preserving IIL performance. Notably, recent model-based imitation methods~\cite{dino_bot, functo} can perform robotic manipulation tasks from one or a few human demonstrations by leveraging generalizable priors from pre-trained models. Although these methods perform competitively in the low-data regime, they are typically surpassed by end-to-end IL policies at larger dataset scales.
% The reason for choosing one-shot rather than few-shot is the former requires less human operation. We pick up some of them to construct a one-shot assistant expert.

Motivated by this observation, this paper introduces Easy-IIL, an IIL framework for robotic manipulation that utilizes off-the-shelf model-based imitation methods as an assistant expert to replace a human expert in the majority of the data collection process in both offline and online stages. Fig. \ref{fig:intro_iil}(b) illustrates the conceptual pipeline of Easy-IIL. Specifically, the human expert only collects a single offline demonstration to initialize the assistant expert and perform intervention when the task is approaching failure. Furthermore, our data collection strategy guarantees that Easy-IIL's performance is comparable to established IIL frameworks by maintaining both offline and online data quality. Simulations and real-world experiments demonstrate that Easy-IIL reduces human operational burden by 9–10 times compared to standard IIL, while maintaining comparable performance. Furthermore, user studies validate that the subjective workload is significantly reduced compared to the baselines.

In brief, the contributions of this paper are organized as follows.
\begin{itemize}
    \item We propose Easy-IIL, an effort-efficient IIL framework designed to minimize human operational effort by utilizing an assistant expert to replace the human in the majority of data collection while maintaining IIL performance.
    
    \item The three key designs of our online interaction strategy—action chunk switching, novice action noising, and bottleneck-space novice action prohibition—contribute critically to preserving IIL performance.
    
    \item Extensive simulations, real-world experiments, and user studies are performed to demonstrate the effectiveness of the proposed Easy-IIL.
\end{itemize}

\section{RELATED WORKS}
\subsection{Interactive Imitation Learning} Many IIL studies alleviate the human monitoring burden by designing real-time anomaly detectors to autonomously decide when intervention is needed, thereby framing the problem as a binary classification task. HG-DAgger \cite{hg_dagger} utilizes the safety metric for this decision, while ThriftyDAgger \cite{thrifty_dagger} and Liu et al. \cite{model_based} employ two metrics: novelty and risk. More recently, Diff-DAgger \cite{diff_dagger} introduces the diffusion loss as the anomaly detection metric in the diffusion policy \cite{dp} to determine when to ask the human expert for help. Our framework complements these methods, aiming to reduce the human's active operational workload rather than the passive monitoring workload.

% Some works focus on dealing with intervention data: IWR \cite{iwr} splits online data into non-intervention and intervention data and re-weight them. Later studies, such as CEILing and Sirius split data into more categories and apply corresponding weights to improve the policy performance. Luo et al. \cite{rlif} reformulated interactive imitation learning as reinforcement learning whose performance with higher upper bound, but requires a large number of interaction data or significant training time. Recently, MILE \cite{mile} employed a mental model to mimic human intervention. However, \cite{mile} require extensive data to train the mental model and prior access to the action distribution of intervention agent, usually human, which are impractical in real-world deployment. In brief, we consider IWR \cite{iwr} and sirius \cite{sirius} as our baseline.

Some existing IIL works focus on enhancing IIL performance. For instance, IWR \cite{iwr} splits online data into non-intervention and intervention subsets and re-weights them. A later study, Sirius \cite{sirius}, further categorizes the online data and applies corresponding weights to improve policy performance. Luo et al. \cite{rlif} reformulate IIL as reinforcement learning (RL), which offers a higher theoretical performance upper bound but demands a large volume of intervention data or significant training time. More recently, MILE \cite{mile} employs a mental model to mimic human intervention. However, MILE requires extensive intervention data to train this mental model and prior access to the action distribution of the intervention agent (typically human), which is impractical for real-world deployment. In brief, these works  \cite{hg_dagger, iwr, sirius} do not focus on how to reduce human's active operational burden.

\subsection{Imitation from Human Demonstrations.}
% These methods are primarily training-free and use pretrained vision model representations and set-up rules to compute the geometric relationship (usually 6D pose) between observed and demonstration objects. Then uses these relationships to transfer the demonstration trajectory and execute it to complete the task \cite{pose_one_shot, ditto}. Many recent works have adopted foundational vision models, such as DINO \cite{dino}, for their representations to further improve the generalization and accuracy \cite{dino_bot, orion, functo}. Notably, while recent work like MimicFunc \cite{mimicfunc} designs an efficient one-shot imitation method to generate offline real-world data for training an imitation learning policy, it does not consider how to interact with novice policy and human expert in the online stage.

Most of these methods are model-based, leveraging rules to utilize pre-trained vision models for computing the spatial object relationship (typically the 6D pose) between the current observation and the demonstration. These relationships are then used to transfer the trajectory in demonstration for task execution \cite{pose_one_shot}. These works adopt pre-trained vision foundation models to further enhance robotic manipulation performance \cite{dino_bot, multi_stage, functo}. Recently, MimicFunc \cite{mimicfunc} introduces efficient strategies to generate offline real-world data for training end-to-end IL policies. We build upon this class of approaches to form our assistant expert.

% To minimize human demonstrations, we adopt one-shot, rather than few-shot, imitation methods as the basic structure of our imitation expert. 

% \begin{figure*}[htbp]
% \centering
% \includegraphics[scale=0.31]{figs/method_5.pdf}
% \caption{``Data Collection Strategy" presents our proposed data collection pipeline covering both offline and online stages. ``Automated Expert" briefly describes the automated expert with the minimal modification termed ``Recovery".}
% \label{fig:method}
% \end{figure*}

\subsection{Data Generation for Imitation Learning}
These studies generate a large number of demonstrations from a few human demonstrations. Methods like MimicGen \cite{mimicgen}, SkillGen \cite{skillgen}, and CyberDemo \cite{cyberdemo} leverage human demonstrations and privileged information (inaccessible in the real world) from simulation to generate demonstrations with diverse initial conditions. However, they focus solely on offline data and suffer from sim-to-real gaps. Recent Real2Sim2Real work \cite{r2r2r} trains IL policies by reconstructing real-world objects and recording their motion in simulation to generate extensive demonstrations. However, this method heavily relies on the quality of 3D reconstruction and object tracking. DemoGen \cite{demogen} requires only one human demonstration to generate demonstrations in the real world. However, it remains limited to offline data generation and only supports point-cloud-input policies. Our work collects both offline and online data directly in the real world, eliminating sim-to-real or reconstruction gaps and supporting policies with diverse types of visual inputs.

\section{METHOD}
\subsection{Overall Workflow}
Easy-IIL begins with an offline phase. Initially, a human expert collects a single demonstration of a manipulation task to initialize both an assistant expert and a dataset. Subsequently, the assistant expert collects the remaining demonstrations and adds them to the dataset, with human intervention occurring only when the task is on the verge of failure. Then, the dataset is used to train an end-to-end imitation learning policy, yielding a novice policy.

This is followed by an online iteration phase. The novice policy is deployed to perform the manipulation task. During this phase, the assistant expert randomly intervenes in the novice policy's execution, while the human intervenes only to prevent imminent failure. The online intervention data from both experts are then incorporated into the dataset to further improve the novice policy.

The pseudo-code for the overall Easy-IIL workflow is provided in Algorithm \ref{alg:easy_iil_part1}. Specifically, \textsc{Activate}$(\cdot)$ refers to the initialization of an assistant expert, while \textsc{Deploy}$(\cdot)$ implements the data collection strategy where the labels \textit{ONE\_DEMO}, \textit{REST\_DEMO}, and \textit{CORRECTION} are used to represent the collection of the first offline demonstration, the remaining offline demonstrations, and the online intervention data respectively. Finally, \textsc{Train}$(\cdot)$ executes the training process.

\begin{figure*}[htbp]
\vspace*{0.325cm}% 将下图整体下移；调大/调小此值可增减下移距离
\centerline{\includegraphics[width=0.92\linewidth]{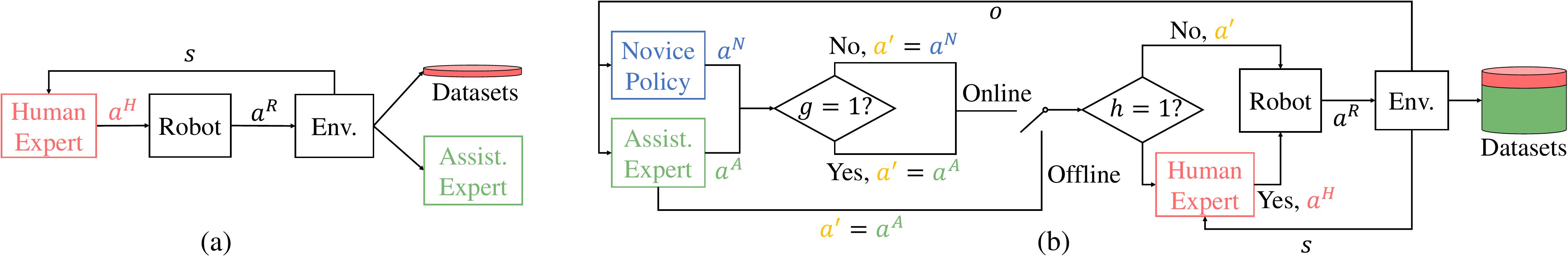}}
\caption{Data collection strategy in Easy-IIL for (a) the first one offline demonstration and (b) the remaining offline demonstrations (``Offline" mode) and online interaction data (``Online" mode). Specifically, $a^H=\pi^H(x)$, $a^R=\pi^R(x, o)$, $a^A=\pi^A(o)$, $a^N=\pi^N(o)$. $x$ and $o$ represent state and observation, respectively.}
\label{fig:method}
\end{figure*}

\subsection{Data Collection Strategy}
% The interactive method resolves two key issues: (1) Action selection: determining whose action to choose at any moment (offline/online), which aligns with the ``Select" module (Fig. 1a). (2) Replanning: When control switches to the assistant expert's action, how does it replan to deal with the deviation caused by the prior novice policy or human actions? The latter is not considered by existing one-shot imitation learning approaches.

% Data collection strategy is the core of Easy-IIL. We first introduce its procedure in offline and online stage and then present three modifications for DAgger.

\subsubsection{Offline Data Collection}
In the offline phase, a human expert provides a single demonstration that initializes an assistant expert and also serves as the initial dataset. Once initialized, the assistant expert acquires the ability to complete the task independently and is utilized to generate the remaining offline demonstrations. The human expert monitors this generation process, intervening only upon task anomalies. Let $x_k$, $o_k$, and $a_k$ denote the state, observation, and action at time step $k$, respectively \cite{e2e_vis_p}. Specifically, the state represents human perception input, whereas the observation corresponds to sensor readings \cite{hg_dagger}. Let $\pi^R_0$ denote the robot policy controlling the robotic arm during the offline stage, defined as: 
\begin{equation}
    \pi^R_0(x_k, o_k) = h(k)\pi^H(x_k) + (1-h(k))\pi^A(o_k)
\end{equation} where $\pi^H$ and $\pi^A$ represent the human expert and assistant expert policies, respectively. The term $h(k)$ serves as a gating function indicating whether the human takes control of the robot at step $k$:
\begin{equation}
h(k)=
\begin{cases}
1, & \text{human intervenes.} \\
0, & \text{otherwise.}
\end{cases}
\label{eq:human_gate_func}
\end{equation}

The offline dataset where data are generated by $\pi^H$ and $\pi^A$ is defined as $\mathcal{D}^{off}=\{(o, a)|a \in \{\pi^H(x), \pi^A(o)\}\}$ and $\mathcal{D}^{off}=\mathcal{D}^{one} \cup \mathcal{D}^{rest}$, where $\mathcal{D}^{one}$ and $\mathcal{D}^{rest}$ are declared in Algorithm \ref{alg:easy_iil_part1}. Upon the completion of data collection, $\mathcal{D}^{off}$ is utilized to train an end-to-end imitation learning policy. This yields a novice policy $\pi^N_0$, which initially exhibits low success rates and produces suboptimal actions. The offline data collection process is illustrated in Fig. \ref{fig:method}(a) and Fig. \ref{fig:method}(b) with ``Offline" mode.

\subsubsection{Online Data Collection}
\begin{figure*}[t]
\centering
\vspace*{0.42cm}% 将下图整体下移；调大/调小此值可增减下移距离
\includegraphics[scale=0.26]{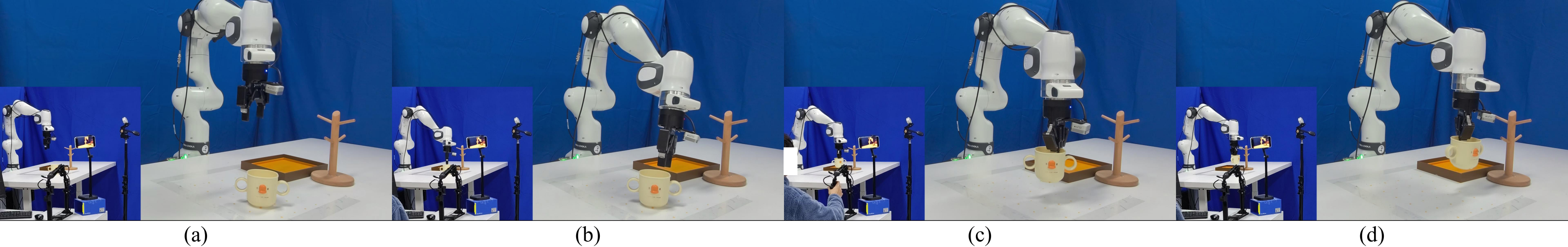}
\caption{Visualization of an online interaction trajectory of Easy-IIL on the ``Hang the Cup" task. The robot arm is controlled by (a) the novice policy; (b) the assistant expert, which yields a skewed grasping pose that risks failure; (c) the human expert, who intervenes and successfully picks up the cup; and (d) the assistant expert, which moves the arm normally toward the cup holder.}
\label{fig:online_procedure_vis}
\end{figure*}

The online stage is divided into $M$ rounds. In the $i$-th round ($i=1,...,M$), a gating function $g_i(j,k)$ selects a sequence of actions (action chunk) from the novice policy $\pi^N_{i-1}$ or the assistant expert $\pi^A$ randomly. Simultaneously, a human expert monitors the execution, providing intervention actions exclusively upon near task failure. Taking the ``Hang the Cup'' task as an example, the visualization of a single online interaction trajectory is shown in Fig. \ref{fig:online_procedure_vis}. The human gating function $h(k)$ has been defined in Eq. (\ref{eq:human_gate_func}). Therefore, the robot policy in the $i$-th round $\pi^R_i$ is defined as follows:
\begin{equation}
\begin{split}
\pi^R_i(x_k, o_k) & = h(k)\pi^H(x_k) + (1-h(k))[g_i(j,k)\pi^A(o_k) \\ & + (1-g_i(j,k))\pi^N_{i-1}(o_k)].
% \hat{\pi}_i(x_k) & = h(k)\pi^H(x_k) + (1-h(k))\tilde{\pi}^N_{i-1}(o_k).
\end{split}
\end{equation}
The definition of $g_i(j,k)$ is shown below,
\begin{equation}
g_i(j,k) = \begin{cases}
1, & X_j < \alpha_i  \ \text{or} \ x_k \in \mathcal{B} \\
0, & \text{otherwise}
\end{cases}
\label{eq:random_switching}
\end{equation}
where $\alpha_i=\beta^i$, with $\beta \in (0,1)$ being defined manually. A larger $\beta$ leads to a higher probability of executing the assistant expert's actions and a slower decay of this probability. At the beginning, $j=0$ and $k=0$. Once an action is executed, $k=k+1$; if an action is executed without human expert intervention, $j=j+1$. Specifically, $X_j$ is defined below,
\begin{equation}
X_j = \begin{cases}
Y_j, & \mathrm{mod}(j,H)=0 \\
X_{j-1}, & \text{otherwise}
\end{cases}
\end{equation}
where $Y_j \overset{\text{i.i.d.}}{\sim} \mathcal{U}(0,1)$ and $\mathrm{mod}(j,H)$ equals the remainder of $j$ divided by $H$, the action-chunk switching size. Compared to single action switching \cite{dagger}, the action chunk switching we used is better suited to end-to-end chunk-based policies, such as diffusion policy and ACT \cite{act}. To enhance exploration for better online action quality, we add zero-mean Gaussian noise with standard deviation $\sigma$ to the normalized novice policy actions with range $[-1, 1]$. Following \cite{multi_stage, skillgen, demogen}, the manipulation workspace is divided into free space $\mathcal{F}$ and bottleneck space $\mathcal{B}$. How this partition is made is detailed in Section~\ref{subsubsec:assist_expert}. $\mathcal{F}$ consists of an open area without obstacles, while $\mathcal{B}$ refers to a region near or in contact with the target object. Since a robot approaching a target object at a closer range requires higher action precision, it is easier to fail in $\mathcal{B}$, especially for novice policy actions due to their low quality. Therefore, we disable the execution of novice policy actions within $\mathcal{B}$, which is implemented by Eq. (\ref{eq:random_switching}) with $x_k \in \mathcal{B}$.

The online dataset where data are generated by $\pi^H$, $\pi^A$, and $\pi^N$ is defined as $\mathcal{D}^{on}=\{(o, a)|a \in \{\pi^H(x), \pi^A(o), \pi^N(o)\}\}$ and $\mathcal{D}^{on}=\mathcal{D}^{corr_1} \cup \mathcal{D}^{corr_2}... \cup \mathcal{D}^{corr_M}$, where $\mathcal{D}^{corr_i} \ (i=1,...,M)$ is declared in Algorithm \ref{alg:easy_iil_part1}. Note that the training weight of online data generated by the novice policy is set to 0, as detailed in Section~\ref{subsubsec:training_details}. After online data collection is completed in the $i$-th round, $\mathcal{D}=\mathcal{D}^{off} \cup \mathcal{D}^{on}$ is used to train $\pi_{i-1}^N$, as shown in Algorithm \ref{alg:easy_iil_part1}. The online data collection process is illustrated in Fig.~\ref{fig:method}(b) with
``Online" mode. The data collection strategy corresponding to \textsc{Deploy}$(\cdot)$ is shown in Algorithm~\ref{alg:easy_iil_part2}.

\begin{algorithm}[t]
\caption{Easy-IIL}
\label{alg:easy_iil_part1}
\small
\begin{algorithmic}
    \Require Rounds $M$, Offline Demonstration Number $N_0$, Demonstration Number in the $i$-th Online Round $N_i$, Data Weights $\mathcal{W}$.
    % \Ensure Optimized Policy $\pi_{\theta}$
    % \State \textbf{Hyperparameters:} Learning rate $\alpha$, Batch size $B$
    \State $\mathcal{D} \leftarrow \emptyset$, $\mathcal{W} \leftarrow \emptyset$.

    \State Label = \textit{ONE\_DEMO} \Comment{One Human Demo}
    % \State $\{\tau_1\}, \{w_1\} \leftarrow \textsc{Deploy}(\pi^H, \text{None}, \text{None}, 1, \text{Label})$
    \State $\mathcal{D}^{one}, \mathcal{W}^{one} \leftarrow \textsc{Deploy}(\pi^H, \text{None}, \text{None}, 1, \text{Label},\text{None})$
    \State $\mathcal{D} \leftarrow \mathcal{D}^{one}$, $\mathcal{W} \leftarrow \mathcal{W}^{one}$
    \State $\pi^A \leftarrow \textsc{Activate}(\mathcal{D})$
    \State Label = \textit{REST\_DEMO}  \Comment{Rest Offline Demos}
    \State $\mathcal{D}^{rest}, \mathcal{W}^{rest} \leftarrow \textsc{Deploy}(\pi^H, \pi^A, \text{None}, N_0-1, \text{Label},\text{None})$
    \State $\mathcal{D} \leftarrow \mathcal{D} \cup \mathcal{D}^{rest}$, $\mathcal{W} \leftarrow \mathcal{W} \cup \mathcal{W}^{rest}$
    \State $\pi^N_0 \leftarrow \textsc{Train}(\mathcal{D}, \mathcal{W}, \text{None})$

    \State Label = \textit{CORRECTION} \Comment{Online Corrections}
    \For{$i=1, 2, \dots, M$}
        \State $\alpha_i \leftarrow \beta^i$
        \State $\mathcal{D}^{corr_i}, \mathcal{W}^{corr_i} \leftarrow \textsc{Deploy}(\pi^H, \pi^A, \pi^N_{i-1}, N_i, \text{Label},\alpha_i)$
        \State $\mathcal{D} \leftarrow \mathcal{D} \cup \mathcal{D}^{corr_i}$, $\mathcal{W} \leftarrow \mathcal{W} \cup \mathcal{W}^{corr_i}$
        \State $\pi^N_i \leftarrow \textsc{Train}(\mathcal{D}, \mathcal{W}, \pi^N_{i-1})$
    \EndFor
    \State \Return $\pi^N_M$
    \algstore{myalg} 
\end{algorithmic}
\end{algorithm}

\subsection{Implementation Details}
\subsubsection{Assistant Expert}\label{subsubsec:assist_expert}
We build upon these works \cite{dino_bot, multi_stage} as our assistant expert to achieve model-based multi-stage robotic manipulation. For initialization, a human expert collects a single offline demonstration where each waypoint consists of an RGB-D image and the corresponding 6D end-effector (EE) pose and 1D gripper value. The human manually segments the demonstration trajectory into stages and further partitions each stage into free-space $\mathcal{F}$ and bottleneck-space $\mathcal{B}$ segments. This partitioning is performed only once per task. After initialization, the assistant expert takes the current RGB-D observation as input. Then, Grounded-SAM \cite{groundedsam}, and image matching techniques (SuperPoint \cite{superpoint} and LightGlue \cite{lightglue}) are employed to process RGB-D images from both the demonstration and the input, thereby extracting a pair of target object point clouds with established correspondences. Subsequently, the 6D pose transformation is computed via Iterative Closest Point (ICP) \cite{icp} and RANSAC \cite{ransac}. Within the demonstration trajectory, the 6D poses are transformed using this transformation, while the gripper values remain unchanged, resulting in the transformed trajectory. After that, the transformed trajectory in $\mathcal{F}$ is re-planned using linear interpolation for positions and Spherical Linear Interpolation (Slerp) \cite{slerp} for orientations, while the gripper values remain the same. The endpoint of this replanned trajectory is the first waypoint of the transformed trajectory within the $\mathcal{B}$ space. The output is the first unvisited waypoint from the replanned transformed trajectory, which spans both the $\mathcal{F}$ and $\mathcal{B}$ spaces.

A recovery behavior is triggered whenever the deviation between the current 6D pose and the replanned transformed trajectory exceeds a predefined threshold. In this case, a new replanned transformed trajectory is generated by the replanning procedure described above, starting from the current 6D pose and targeting the first unvisited waypoint within $\mathcal{B}$.

\subsubsection{Training Details}\label{subsubsec:training_details}
We employ the diffusion policy \cite{dp} as our end-to-end chunk-based IL policy. The training process utilizes the DDPM noise scheduler \cite{ddpm}, with Mean Squared Error (MSE) serving as the loss function. The training loss \cite{dp} is formulated as follows,
\begin{equation}
    \mathcal{L} := \mathbb{E}_{a^0_{k:k+H^p-1}, o_k, n, \epsilon} [\left\| \epsilon - \epsilon_\theta(a^n_{k:k+H^p-1}, o_k, n) \right\|_2^2]
\end{equation}
where $\epsilon \sim \mathcal{N}(\mathbf{0},\mathbf{1})$ is the standard Gaussian noise, $\epsilon_\theta (\cdot)$ is the noise prediction model, $a^0_{k:k+H^p-1}$ represents the clean action chunk, $a^n_{k:k+H^p-1}$ represents the noisy action chunk at the $n$-th scheduling step $(n>0)$, and $H^p$ means the action chunk size for prediction.

We define a weight vector $w_k \in \mathbb{R}^{H^p}$, whose component $[w_k]_j$ ($j \in \{0,\dots,H^p-1\}$) weights the action at time step $k+j$ within the chunk at step $k$ during training. Each component is 1 by default. If the robot is controlled by the novice policy at step $k$, the entire vector is zeroed, i.e., $w_k=\mathbf{0}$; otherwise, for each $j \in \{1,\dots,H^p-1\}$, we set $[w_k]_j=0$ whenever the robot is controlled by the novice policy at step $k+j$. Consequently, the weighted training loss is formulated as:
\begin{equation}
    \mathcal{L} := \mathbb{E}_{a^0_{k:k+H^p-1}, o_k, n, \epsilon} [\left\| w_k 
\otimes (\epsilon - \epsilon_\theta(a^n_{k:k+H^p-1}, o_k, n)) \right\|_2^2].
\label{eq:weightedDDPMLoss}
\end{equation}

The detailed training hyperparameters are provided in Table \ref{tab:hyperparameters} in the Appendix.

% This weighting trick is applied to the diffusion policy in all subsequent experiments.

% To ensure both offline and online training stability, we incorporate Exponential Moving Average (EMA) and a cosine learning rate (LR) scheduler with warmup \cite{cosine_lr, warmup}. We initialize both the EMA and the cosine LR scheduler at the start of the offline stage. For the online stage, the EMA is initialized once at the beginning, but the cosine LR scheduler without warmup is re-initialized at the start of every new round. For the chunk-based imitation learning policy, we assign a weight to each action within the chunk based on its corresponding state. This weighting trick is applied to the diffusion policy in all subsequent experiments.

\section{EXPERIMENTS}
Our experiments are designed to answer the following questions: (1) Can Easy-IIL significantly reduce the human operational workload while maintaining IL policy performance? (2) What is the contribution of each module of Easy-IIL to the overall performance? (3) As the amount of data increases, can the end-to-end IL policy trained with Easy-IIL surpass the rule-based assistant expert? (4) Can the assistant expert be replaced by pretrained vision-language-action (VLA) models \cite{pi-0.5} with strong generalization capabilities?

\begin{figure}[t]
\centering
\vspace*{0.3cm}
\includegraphics[scale=0.082]{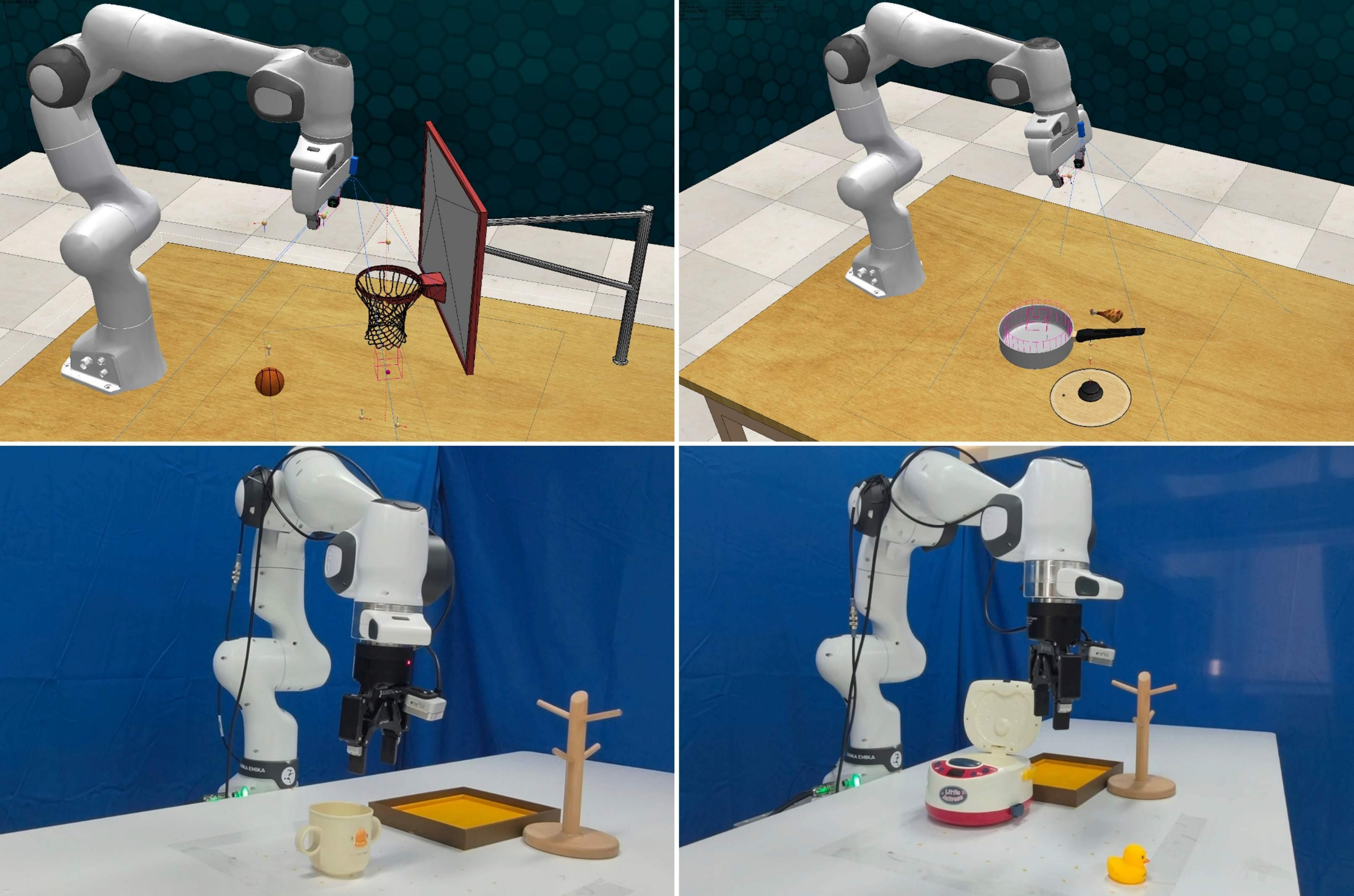}
\caption{Four task scenarios. Basketball in Hoop (left-top), Take Chicken in Saucepan (right-top), Hang the Cup (left-bottom), and Put Duck in Cooker (right-bottom).}
\label{fig:task_scene}
\end{figure}

\subsection{Tasks}
\textbf{Simulations:} Our simulations are conducted in RLBench \cite{rlbench}, featuring both an existing two-stage task and a custom long-horizon four-stage task. The two-stage task, ``Basketball in Hoop", involves picking up a basketball and placing it in a hoop. The custom-designed four-stage task, ``Take Chicken in Saucepan", requires picking up a chicken, placing it in a saucepan, retrieving the lid, and covering the pan. The simulated task scenes are shown in the top two panels of Fig. \ref{fig:task_scene}. A human controls the four degrees of freedom (3D position and 1D gripper) of the end-effector via a keyboard.

\textbf{Real-World Experiments:} We set up two manipulation tasks. The first is a two-stage task named ``Hang the Cup", which involves grasping a cup and hanging it on a holder. The second is a three-stage task, ``Put Duck in Cooker", which requires picking up a toy duck, placing it into a cooker, and closing the cooker. The real-world task scenes are illustrated in the bottom two panels of Fig. \ref{fig:task_scene}.

The real-world experiments utilize a 7-DoF Franka Research 3 robotic arm, with the Gello \cite{gello} serving as a teleoperation interface. A human expert teleoperates the robotic arm by controlling the Gello, facilitated by a direct joint-to-joint mapping.

\begin{figure*}[htbp]
\centering
\vspace*{0.3cm}
\includegraphics[scale=0.51]{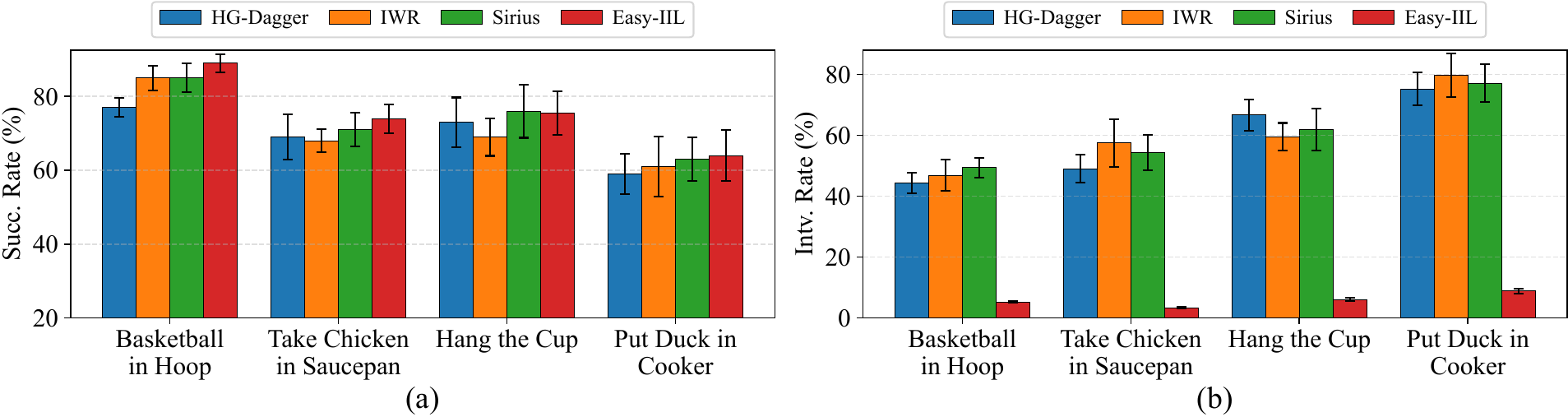}
\caption{Main experiment results evaluating Succ. Rate (a) and Intv. Rate (b) across four tasks for Easy-IIL, HG-DAgger, IWR, and Sirius. Easy-IIL is configured with $H=8$ and $\sigma=0.3$.}
\label{fig:main_result_test}
\end{figure*}

\subsection{Baselines and Metrics}
\subsubsection{Baselines} Three mainstream IIL methods, HG-DAgger, IWR, and Sirius, are employed as baselines. These baselines follow the standard IIL framework (introduced in Section~\ref{sec:intro}) for data collection. We tune the hyperparameters of HG-DAgger, IWR, and Sirius to ensure a fair comparison. These three methods are collectively referred to as IIL Baselines.

\subsubsection{Metrics} For the main, ablation, and additional experiments, we evaluate the IIL performance using the task Success Rate (Succ. Rate), defined as the rate at which the trained end-to-end policy successfully executes the assigned task. Also, we evaluate human operational burden by human intervention rate (Intv. Rate). Intv. Rate is a variant of the normalized Return On Human Effort (ROHE) \cite{model_based, robo_fleet}, modified to retain only the denominator term, thereby allowing for the direct quantification of human operational burden. The intervention rate is expressed as follows:
% \begin{equation}
% \text{Intv. Rate} = \frac{N^H}{N}
% \end{equation}
\begin{equation}
\text{Intv. Rate} = \frac{N^H}{N} \times 100\%
\end{equation}
where $N^H$ and $N$ denote the number of human actions and the total number of actions, respectively, in the dataset of a given task.
% The configuration concerning the checkpoint evaluation schedule for simulations and real-world experiments is provided in the Table III in Appendix.A.

For the user study, we employ the Raw-TLX \cite{raw_tlx} to quantify the subjective operational workload of participants during interactive imitation learning across six metrics: mental demand, physical demand, temporal demand, performance, effort, and frustration. Each metric is scored on a 0--100 scale with an interval of 5, where a lower score indicates a lower workload or better performance.

The Raw TLX score is the mean of these six metric scores and serves as the final indicator for quantifying the subjective operational workload of participants. We employ the Wilcoxon Signed-Rank Test to conduct a significance test on the Raw TLX score between Easy-IIL and IIL baselines.

% \begin{figure*}[htbp]
% \centering
% \includegraphics[scale=0.4]{figs/main_result_test.pdf}
% \caption{Main experiment results evaluating Succ. Rate (a) and Intv. Rate (b) across four tasks for Easy-IIL, HG-DAgger, IWR, and Sirius. Easy-IIL is configured with $H=8$ and $\sigma=0.3$.}
% \label{fig:main_result_test}
% \end{figure*}

\subsection{Experiment Details}\label{subsec:exp_details}
The main experiment, ablation studies, the additional experiment, and the user study require 30 demonstrations: 10 offline demonstrations and 4 online rounds, with 5 interactive trajectories per round. In addition, the additional experiment also considers settings with more demonstrations: the 50-demonstration and 80-demonstration settings correspond to 10/20 offline demonstrations and 4/6 online rounds, respectively, with 10 interactive trajectories per round. We fine-tune a representative VLA model \(\pi_{0.5}\) \cite{pi-0.5} using one and five offline demonstrations in the additional experiment. Its language instructions for the ``Basketball" and ``Chicken" tasks are ``Put the basketball in the hoop" and ``Put the chicken leg in the saucepan and cover it with the lid", respectively.

The main experiment comprises two simulated tasks performed by 10 participants (aged 20–27) using a fixed random seed, along with two real-world tasks conducted by five participants randomly selected from the 10 participants over three trials. The ablation and additional experiments focus on two simulated tasks, involving a single participant across three different random seeds. For the user study, the 10 participants rate the two simulation tasks and the five participants rate the two real-world tasks, based on the entire operation process. The order of the methods used by each participant is randomized.

In the simulation, we evaluate 60 trajectories per checkpoint for each seed to compute the success rate, and for each checkpoint, we average the success rate across all seeds. We then select the highest mean among the six checkpoints as the reported success rate. For real-world experiments, we evaluate 30 trajectories per trial (using the last checkpoint for each trial) and report the mean success rate across all trials.

Note that ``Basketball", ``Chicken", ``Cup", and ``Duck" refer to the Basketball in Hoop, Take Chicken in Saucepan, Hang the Cup, and Put Duck in Cooker tasks, respectively, in Tables \ref{tab:off_human_vs_easyiil}, \ref{tab:hyperparameter_ablation}, \ref{tab:add_exp1}, and \ref{tab:tlx_results}.

\begin{figure}[htbp]
\centering
\includegraphics[scale=0.45]{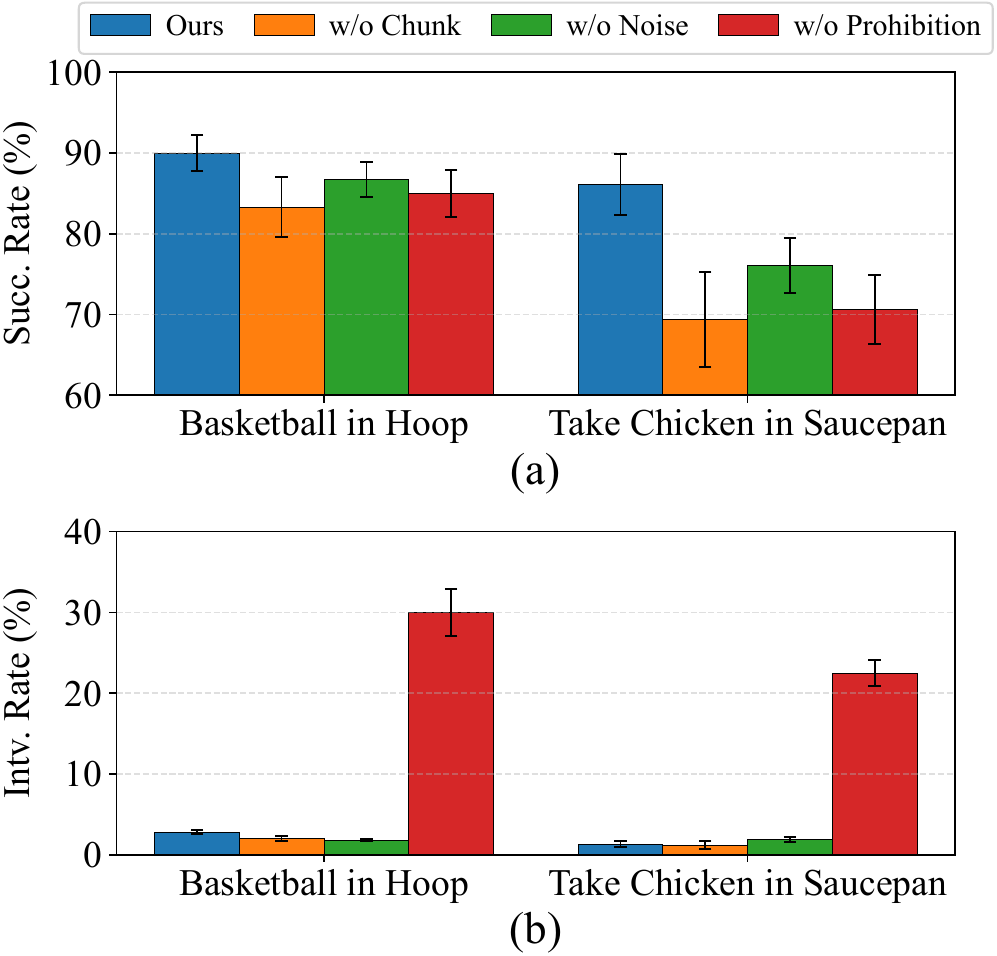}
\caption{Ablation studies results evaluating (a) Succ. Rate and (b) Intv. Rate across two simulated tasks for online strategy of Easy-IIL. Easy-IIL is configured with $H=8$ and $\sigma=0.3$.}
\label{fig:ablation_result1}
\end{figure}

\subsection{Experiment Results}
\subsubsection{Main Results}
Fig. \ref{fig:main_result_test}(a) demonstrates that Easy-IIL consistently achieves performance comparable to all IIL baselines, with average success rate differences less than 4\% and the standard deviation differences within 3\%. As shown in Fig. \ref{fig:main_result_test}(b), Easy-IIL significantly reduces the human intervention rate across all tasks, achieving a substantial nine- to ten-fold reduction. These results verify the effectiveness of Easy-IIL.

\subsubsection{Ablation Results}
As illustrated in Table \ref{tab:off_human_vs_easyiil}, across the two simulated tasks, the offline Succ. Rate of Easy-IIL is comparable to that of the baselines. The offline Intv. Rate of Easy-IIL is substantially lower than that of baselines. These results confirm the effectiveness of Easy-IIL in the offline stage. Note that the best-performing baseline (the one with the highest sum of success rates across the two simulated tasks) is used as the ``Best Baseline" in Table \ref{tab:off_human_vs_easyiil}.

\begin{table}[h]
\centering
\caption{Ablation study results evaluating the offline Succ. Rate and Intv. Rate across two simulated tasks for the best baseline and Easy-IIL. Results are presented as Mean $\pm$ Standard Deviation.}
\label{tab:off_human_vs_easyiil}

\resizebox{0.49\textwidth}{!}{
\begin{tabular}{ccccc}
\toprule

& \multicolumn{2}{c}{Succ. Rate (\%)} & \multicolumn{2}{c}{Intv. Rate (\%)} \\
\cmidrule(lr){2-3} \cmidrule(lr){4-5}

Method & Basketball & Chicken & Basketball & Chicken \\
\midrule

Best Baseline & 74.4 $\pm$ 2.1 & 68.3 $\pm$ 5.3 & 100.0 $\pm$ 0.0 & 100.0 $\pm$ 0.0 \\
Easy-IIL      & 77.8 $\pm$ 2.6 & 65.0 $\pm$ 4.7 & 4.6 $\pm$ 0.1 & 2.8 $\pm$ 0.2 \\
\bottomrule
\end{tabular}
}
\end{table}

Fig. \ref{fig:ablation_result1}(a) illustrates that for both simulation tasks, incorporating action chunk switching (Ours vs. w/o Chunk), injecting noise into novice policy actions (Ours vs. w/o Noise), and disabling novice actions in $\mathcal{B}$ (Ours vs. w/o Prohibition) contribute to a higher Succ. Rate. Additionally, Fig. \ref{fig:ablation_result1}(b) shows that disabling novice actions in $\mathcal{B}$ significantly reduces human online intervention across both tasks. These results confirm the effectiveness of Easy-IIL in the online stage.

% Also, we investigate the impact of different values for action chunk switching size $H$ and novice policy action noise standard deviation $\sigma$ on the final results to identify the optimal values. Upper sub-table in Table \uppercase\expandafter{\romannumeral1} shows that the policy achieves the best performance across both tasks when $H=H^{pred}$. Lower sub-table in Table \uppercase\expandafter{\romannumeral1} confirms that the optimal performance is reached when $\sigma=0.3$.

We investigate the sensitivity of IIL performance to the variations of parameters ($H$ and $\sigma$) in Easy-IIL's online strategy, as detailed in Table \ref{tab:hyperparameter_ablation}. It can be observed that the average Succ. Rate remains stable within a 3\% margin and the standard deviations stay within 4\% in the vast majority of cases as parameters vary, underscoring the robustness of Easy-IIL's online strategy against parameter variations.

% Furthermore, we quantitatively investigate the impact of the action chunk size $H$ and the action noise standard deviation $\sigma$ on the final policy performance. As shown in the upper section of Table \uppercase\expandafter{\romannumeral1}, the policy achieves peak performance across both tasks when $H=H^{pred}$. Similarly, the lower section of Table \ref{tab:hyperparameter_ablation} confirms that optimal performance is attained at $\sigma=0.3$. Additionally, Easy-IIL demonstrates robustness to $H$ and $\sigma$, as the success rate remains stable within a 3\% margin.

\begin{table}[htbp]
    \centering

    \caption{Ablation study results evaluating Succ. Rate across two simulated tasks for parameter variations in the online strategy of Easy-IIL. Top: varying $H$ (with $\sigma=0.3$); Bottom: varying $\sigma$ (with $H=8$). Results are presented as Mean $\pm$ Standard Deviation.}

    \resizebox{0.4\textwidth}{!}{%
    \begin{tabular}{lccc}
      \toprule
      Succ. Rate (\%) & $H=4$ & $H=8$ & $H=12$ \\
      \midrule
      Basketball & 90.0 $\pm$ 3.1 & 90.0 $\pm$ 2.2 & 88.9 $\pm$ 1.9 \\
      Chicken    & 84.4 $\pm$ 3.5 & 86.1 $\pm$ 3.8 & 83.3 $\pm$ 5.1 \\
      \bottomrule
    \end{tabular}
    }
    
    \vspace{0.2cm}

    \resizebox{0.4\textwidth}{!}{%
    \begin{tabular}{lccc}
      \toprule
      Succ. Rate (\%) & $\sigma=0.2$ & $\sigma=0.3$ & $\sigma=0.4$ \\
      \midrule
      Basketball & 88.3 $\pm$ 2.1 & 90.0 $\pm$ 2.2 & 88.9 $\pm$ 3.8 \\
      Chicken    & 84.4 $\pm$ 4.3 & 86.1 $\pm$ 3.8 & 83.9 $\pm$ 3.7 \\
      \bottomrule
    \end{tabular}
    }
    \label{tab:hyperparameter_ablation}
\end{table}

% is possible from a good trade-off between exploration and exploitation, along with the compatibility with action-chunk-based policies. 
% The latter value was determined with actions normalized to the range from -1 to 1, suggesting its invariance.

% Furthermore, Table \ref{tab:add_exp_weight} ablates the training data weighting strategy in Easy-IIL. ``Ours'' achieves the highest Succ. Rate across both simulated tasks, validating the design of assigning zero weight to novice-policy actions and retaining only samples whose first action originates from an expert.

% \begin{table}[htbp]
%     \centering
%     \caption{Ablation study results evaluating Succ. Rate across two simulated tasks for different training data weighting strategies in Easy-IIL. ``w/o Weighting'' assigns a weight of 1 to all actions; ``Ours'' assigns weight 0 to novice-policy actions and weight 1 to expert actions while retaining only samples whose first action originates from an expert; ``w/o Selection'' adopts the same 0/1 weighting but retains all samples. Results are presented as Mean $\pm$ Standard Deviation.}
%     \label{tab:add_exp_weight}

%     \resizebox{0.48\textwidth}{!}{%
%     \begin{tabular}{lccc}
%       \toprule
%       Succ. Rate (\%) & w/o Weighting & Ours & w/o Selection \\
%       \midrule
%       Basketball & 84.7 $\pm$ 3.2 & \textbf{90.0 $\pm$ 2.2} & 87.6 $\pm$ 4.1 \\
%       Chicken    & 81.4 $\pm$ 5.6 & \textbf{86.1 $\pm$ 3.8} & 83.2 $\pm$ 4.3 \\
%       \bottomrule
%     \end{tabular}
%     }
% \end{table}

\subsubsection{Additional Results}
Table \ref{tab:add_exp1} shows that as the amount of data increases, the end-to-end IL policy outperforms the model-based assistant expert with handcrafted rules, with the standard deviation differences within 2\%. Note that assignments of 30, 50 and 80 demonstrations are mentioned in the first paragraph of Section~\ref{subsec:exp_details}.

\begin{table}[htbp]
    \centering
    \vspace*{0.3cm}
    \caption{The additional experiment evaluating Succ. Rate for assistant expert (Assist. Expert) and diffusion policy trained via Easy-IIL with 30 (DP-30), 50 (DP-50), and 80 demonstrations (DP-80). Results are presented as Mean $\pm$ Standard Deviation.}

    \resizebox{0.48\textwidth}{!}{%
    \begin{tabular}{lcccc}
      \toprule
      Succ. Rate (\%) & Assist. Expert & DP-30 & DP-50 & DP-80 \\
      \midrule
      Basketball & 91.7 $\pm$ 2.1 & 90.0 $\pm$ 2.2 & 95.6 $\pm$ 2.5 & 97.8 $\pm$ 1.9 \\
      Chicken    & 88.9 $\pm$ 3.5 & 86.1 $\pm$ 3.8 & 90.6 $\pm$ 3.4 & 94.4 $\pm$ 3.9 \\
      \bottomrule
    \end{tabular}
    }
    \label{tab:add_exp1}
\end{table}

Table~\ref{tab:add_exp2} shows the assistant expert achieves significantly higher success rates than $\pi_{0.5}$ fine-tuned on one or five demonstrations, highlighting its data efficiency and supporting our choice of this expert over a VLA model.

\begin{table}[htbp]
    \centering
    \caption{Success rates of the assistant expert (Assist. Expert) and $\pi_{0.5}$ fully fine-tuned on one ($\pi_{0.5}$-1) and five ($\pi_{0.5}$-5) offline demonstrations. Results are presented as Mean $\pm$ Standard Deviation.}
    \label{tab:add_exp2}
    \resizebox{0.48\textwidth}{!}{%
    \begin{tabular}{lccc}
        \toprule
        Succ. Rate (\%) & Assist. Expert & $\pi_{0.5}$-1 & $\pi_{0.5}$-5 \\
        \midrule
        Basketball & $91.7 \pm 2.1$ & 21.7 $\pm$ 5.1 & 59.4 $\pm$ 2.8 \\
        Chicken    & $88.9 \pm 3.5$ & 14.4 $\pm$ 3.4 & 47.8 $\pm$ 7.2 \\
        \bottomrule
    \end{tabular}%
    }
\end{table}

\subsubsection{User Study Results}
% [Design Description] 
% Copy this paragraph into your "User Study" or "Methodology" section.

% To evaluate the subjective workload and user experience regarding the operation during IIL, we employ the Raw-TLX (unweighted NASA-TLX \cite{nasa_tlx, raw_tlx}). To simplify it, we retain three key metrics: Performance, Effort, and Frustration. 

% 10 participants rate based on the whole process of two simulation tasks. Participants rate each dimension on a 7-point Likert scale (1 = Lowest, 7 = Highest). The definitions for these metrics are established as follows:

% \begin{itemize}
%     \item \textbf{Performance (P):} Reflects the user's satisfaction with the operation results. A higher score indicates a higher level of satisfaction and perceived success.
%     \item \textbf{Effort (E):} Measures the amount of mental and physical effort exerted by the user during operating the robotic arm.
%     \item \textbf{Frustration (F):} Assesses the level of impatience, insecurity, or annoyance experienced during operation.
% \end{itemize}

% The merged score based on the three metrics is defined as
% \begin{equation}
%     \textit{Burden} = \frac{(7-P)+E+F}{3}.
% \end{equation}
% We employed the Wilcoxon Signed-Rank Test \cite{conover1999practical} to conduct a significance test on the subjective operational burden between the two methods.

Table \ref{tab:tlx_results} shows that Easy-IIL achieves lower Raw TLX scores than the IIL baselines across all four tasks, with $p<0.01$, indicating a statistically significant reduction in subjective operational workload. Note that the best-performing IIL baseline (the one with the lowest sum of Raw TLX scores across the four tasks) is used as the ``Best Baseline" in Table \ref{tab:tlx_results}.
% [Table Description]
% Copy the following table block into your "Results" section.
% Note: Replace the placeholder numbers (Mean +/- SD) with your actual data.
% If you only have one group, remove the extra columns.

\begin{table}[htbp]
    \centering
    \caption{User study results evaluating Raw TLX scores for the best baseline and Easy-IIL. Raw TLX scores are presented as Mean $\pm$ Standard Deviation. ($p<0.01$, Wilcoxon Signed-Rank Test).}

    \label{tab:tlx_results}
    \resizebox{0.48\textwidth}{!}{%
    \begin{tabular}{lcccc}
        \toprule
        Raw TLX scores & Basketball & Chicken & Cup & Duck \\
        \midrule
        Best Baseline & $56.5 \pm 10.9$ & $70.3 \pm 12.0$ & $62.5 \pm 8.8$ & $81.7 \pm 15.6$ \\
        Easy-IIL & $25.3 \pm 11.9$ & $33.4 \pm 13.3$ & $28.8 \pm 10.7$ & $38.3 \pm 14.9$ \\
        \bottomrule
    \end{tabular}
    }
    % \vspace{1ex}
    % \\
    % \footnotesize{\textit{Note: $\uparrow$ indicates that a higher score is better; $\downarrow$ indicates that a lower score is better.}}
\end{table}

% \section{DISCUSSION AND CONCLUSIONS}
% The key to Easy-IIL's substantial reduction of human operational burden is that, most of the data produced by the assistant expert is of sufficiently high quality to replace human data for training the IL policy. However, this reliance on model-based methods also introduces a limitation: our framework is restricted to manipulation tasks where model-based methods are effective, i.e., rigid-body or articulated-object manipulation, and does not readily extend to deformable objects. In future work, we plan to transfer our online data collection strategy to a sim-to-real framework, where the assistant expert provides online interventions to the IL policy in simulation to improve simulation data efficiency.

% We propose Easy-IIL, an IIL framework that significantly reduces the human operational burden while maintaining comparable performance. Easy-IIL leverages an off-the-shelf assistant expert to replace the majority of human operations during data collection. The human expert is only required to collect a single offline demonstration to initialize the assistant expert and to intervene in cases of near-failure. Easy-IIL preserves both offline and online data quality to keep IIL performance. Extensive simulation and real-world experiments demonstrate the effectiveness of Easy-IIL. Additionally, a user study confirms that Easy-IIL substantially lowers subjective operational workload.

\section{CONCLUSIONS}
We propose Easy-IIL, an IIL framework that significantly reduces the human operational burden while maintaining performance comparable to IIL baselines. It leverages an off-the-shelf assistant expert to replace the majority of human operations during data collection, requiring the human expert only to collect a single offline demonstration for initializing the assistant expert and to intervene in near-failure cases. By preserving both offline and online data quality, Easy-IIL sustains IIL performance. Extensive simulation and real-world experiments demonstrate its effectiveness, and a user study further confirms that it substantially lowers the subjective operational workload.

\bibliographystyle{IEEEtran}
\bibliography{ref}

\section*{APPENDIX}
\subsection{Hyperparameter Settings}
The hyperparameter setting for diffusion policy in the simulation is shown in Table \ref{tab:hyperparameters}. For real-world experiments, we use the same hyperparameter settings as in simulation, except that the action output is changed to joint angles. $H^p$, $H^{exec}$, and $H^{obs}$ denote
the prediction, execution, and observation horizons of the diffusion
policy, respectively.

\begin{table}[h]
\centering
\vspace*{0.3cm}
\caption{Parameter Setting}
% \begin{tabular}{p{4cm}|p{2cm}}
\small
\begin{tabular}{cc}
\toprule
Parameter & Value \\
\midrule
Image type & Wrist \\
Proprioceptive input & Joint angle \\
Action output & Delta EE pose \\
Image encoder & ResNet-18 \\
Optimizer & AdamW \\
Learning rate & 1e-4 \\
Weight decay & 1e-6 \\
Batch size & 64 \\
$H^p$/$H^{exec}$/$H^{obs}$ & 8/4/2 \\
Diff. step embed. dim. & 64 \\
U-net dims. & [64, 128, 256] \\
$H$/$\sigma$/$\beta$ & 8/0.3/0.7 \\
\bottomrule
\end{tabular}
\label{tab:hyperparameters}
\end{table}

\subsection{Algorithm of Data Collection Strategy}
\noindent\begin{minipage}{\linewidth}
\par\noindent\hrulefill\par\vspace{1pt}
\captionof{algorithm}{Algorithm of Data Collection Strategy}
\label{alg:easy_iil_part2}
\vspace{1pt}\hrule\vspace{3pt}
\small
\begin{algorithmic}
    \algrestore{myalg}

    \Function{Deploy}{$\pi^H, \pi^A, \pi^N, \text{Num}, \text{Label}, \alpha$}
        \State $\mathcal{D}' \leftarrow \emptyset$, $\mathcal{W}' \leftarrow \emptyset$
        \For{$i = 1$ to $\text{Num}$}
            \State Init. env. as $o_0$, $x_0$, $\tau_i \leftarrow \emptyset$, $w^\tau_i \leftarrow \emptyset$, $j \leftarrow 0$, $k \leftarrow 0$,
            \State $Done \leftarrow False$
            \While{$Not \ Done$}
                \begingroup
                \algtext*{EndIf}
                \If{$\mathrm{mod}(j,H) = 0$} $X_j \sim \mathcal{U}(0,1)$
                \Else\ $X_j \leftarrow X_{j-1}$
                \EndIf
                \endgroup

                \begingroup
                \algtext*{EndIf}
                \If{Human intervenes} $h(k) \leftarrow 1$
                \Else\ $h(k) \leftarrow 0$
                \EndIf
                \endgroup

                \begingroup
                \algtext*{EndIf}
                \If{$X_j < \alpha$ or $x_k \in \mathcal{B}$} $g(j,k) \leftarrow 1$
                \Else\ $g(j,k) \leftarrow 0$
                \EndIf
                \endgroup

                \If{Label==\textit{ONE\_DEMO}}
                    \State $a_k \leftarrow \pi^H(x_k)$
                \ElsIf{Label==\textit{REST\_DEMO}}
                    \begingroup
                    \algtext*{EndIf}
                    \If{$h(k)==1$} $a_k \leftarrow \pi^H(x_k)$
                    \Else\ $a_k \leftarrow \pi^A(o_k)$
                    \EndIf
                    \endgroup
                \ElsIf{Label==\textit{CORRECTION}}
                    \If{$h(k)==1$}
                        \State $a_k \leftarrow \pi^H(x_k)$
                    \ElsIf{$g(j,k)==1$}
                        \State $a_k \leftarrow \pi^A(o_k)$
                        \State $j \leftarrow j+1$
                    \Else
                        \State $a_k \leftarrow \pi^N(o_k)$
                        \State $j \leftarrow j+1$
                    \EndIf
                \EndIf 
                
                \State $\tau_i \leftarrow \tau_i \cup \{(o_k,a_k)\}$
                \algtext*{EndIf}
                \If{$a_k$ is from $\pi^H(x_k)$ or $\pi^A(o_k)$} $w^\tau_i \leftarrow w^\tau_i \cup \{1\}$
                \Else\ $w^\tau_i \leftarrow w^\tau_i \cup \{0\}$
                \EndIf
                \algtext{EndIf}{\algorithmicend\ \algorithmicif}
                \State $o_{k+1}, x_{k+1}, Done \leftarrow \text{env.step}(a_k)$
                \State $k \leftarrow k+1$
            \EndWhile 
            
            \State $\mathcal{D}' \leftarrow \mathcal{D}' \cup \{\tau_i\}$, $\mathcal{W}' \leftarrow \mathcal{W}' \cup \{w^\tau_i\}$ 
        \EndFor 
        \State \Return $\mathcal{D}'$, $\mathcal{W}'$
    \EndFunction
\end{algorithmic}
\vspace{1pt}\hrule
\end{minipage}

\end{document}